\documentclass{article}

\usepackage{arxiv}

\usepackage[utf8]{inputenc} 
\usepackage[T1]{fontenc}    
\usepackage{hyperref}       
\usepackage{url}            
\usepackage{booktabs}       
\usepackage{amsfonts}       
\usepackage{nicefrac}       
\usepackage{microtype}      
\usepackage{lipsum}
\usepackage{graphicx}
\usepackage{array}
\graphicspath{ {./images/} }
\usepackage{tikz}

\newcommand\copyrighttext{
  \footnotesize \textcopyright 2024 IEEE. Personal use of this material is permitted. Permission from IEEE must be obtained for all other uses, in any current or future media, including reprinting/republishing this material for advertising or promotional purposes, creating new collective works, for resale or redistribution to servers or lists, or reuse of any copyrighted component of this work in other works.
  
 DOI: \href{https://doi.org/10.1109/RO-MAN60168.2024.10731170}{10.1109/RO-MAN60168.2024.10731170}
  }
\newcommand\copyrightnotice{
\begin{tikzpicture}[remember picture,overlay]
\node[anchor=south,yshift=15pt] at (current page.south) {\fbox{\parbox{\dimexpr\textwidth-\fboxsep-\fboxrule\relax}{\copyrighttext}}};
\end{tikzpicture}%
}

\title{Extended Reality for Enhanced Human-Robot Collaboration: \\ a Human-in-the-Loop Approach}

\author{
    Yehor Karpichev \\
        \texttt{ykarpichev@uvic.ca} \\
    \And
    Todd Charter \\
        \texttt{toddch@uvic.ca} \\
    \And
    Jayden Hong \\
        \texttt{jaydenh@uvic.ca} \\
    \And
    Amir M. Soufi Enayati \\
        \texttt{amsoufi@uvic.ca} \\
    \And
    Homayoun Honari \\
        \texttt{hmnhonari@uvic.ca} \\
    \And
    Mehran Ghafarian Tamizi \\
        \texttt{mehranght@uvic.ca} \\
    \And
    Homayoun Najjaran\thanks{Corresponding author} \\
        \texttt{najjaran@uvic.ca} \\
    \AND
    \\
    Advanced Control and Intelligent Systems Laboratory (ACIS), \\
    University of Victoria, Victoria, BC V8P 5C2, Canada
    \\
}

\begin{document}

\maketitle
\copyrightnotice

\begin{abstract}
The rise of automation has provided an opportunity to achieve higher efficiency in manufacturing processes, yet it often compromises the flexibility required to promptly respond to evolving market needs and meet the demand for customization. Human-robot collaboration attempts to tackle these challenges by combining the strength and precision of machines with human ingenuity and perceptual understanding. In this paper, we conceptualize and propose an implementation framework for an autonomous, machine learning-based manipulator that incorporates human-in-the-loop principles and leverages Extended Reality (XR) to facilitate intuitive communication and programming between humans and robots. Furthermore, the conceptual framework foresees human involvement directly in the robot learning process, resulting in higher adaptability and task generalization. The paper highlights key technologies enabling the proposed framework, emphasizing the importance of developing the digital ecosystem as a whole. Additionally, we review the existent implementation approaches of XR in human-robot collaboration, showcasing diverse perspectives and methodologies. The challenges and future outlooks are discussed, delving into the major obstacles and potential research avenues of XR for more natural human-robot interaction and integration in the industrial landscape.
\end{abstract}

\section{INTRODUCTION}
\label{sec:introduction}
The transition from Industry 4.0, which has facilitated the development and implementation of autonomous cyber-physical systems, IoT, and Big Data in manufacturing, to Industry 5.0 which aims to complement the technological advancements by prioritizing human-centric approaches, fundamentally reshapes the interaction between humans and machines within the manufacturing sector. This evolution involves combining the precision of robots and machines with the intelligence and versatility of human input \cite{jahanmahin2022human}. In this case, one of the most important challenges is the design of communication interfaces that accurately represent the manufacturing processes, account for needs in adaptability and flexibility, and provide intuitive interaction methods for the users. The primary focus of most robot software lies in programming specific functions, with limited to no tolerance for deviations from the programmed settings. Machine learning (ML) offers new possibilities by enhancing the generalization capabilities of robots' decision-making. Nevertheless, it is crucial to recognize ML as an enabling tool rather than as the sole medium for human-robot communication.

Within the conceptual principles of Industry 5.0, human-robot collaboration (HRC) aims to find solutions enabling humans and robots to work together side-by-side, engaging on both physical and cognitive levels. Additionally, the recent advancements in extended reality (XR) technology, encompassing both hardware and software, as well as its integration in digital twins, shows a promising solution to support human involvement as an active agent for HRC purposes \cite{pizzagalli2021ucd} or within the broader context of the smart factory concept. In this paper, the term XR is used as an umbrella term to describe the spectrum comprising augmented, virtual, and mixed reality. Constructing a fully virtual world, Virtual reality (VR) enables users to interact, communicate, sense, and observe virtual objects through a fully immersive experience \cite{dianatfar2021review}. On the other hand, augmented reality (AR) offers a symbiosis of virtual and real by infusing visual augmentations to the physical objects within the existent environment \cite{suzuki2022augmented}. Mixed reality (MR) integrates physical and digital environments, enabling digital visualization overlaid on the physical world and fostering heightened interaction between physical objects and digital interfaces, with some experts considering MR as an advanced extension of AR \cite{speicher2019mixed}. Overall, XR technologies offer a range of human-interaction interfaces tailored for both digital and physical environments, proving advantageous in HRC scenarios where digital and physical elements interact on multiple levels.

The complexity of HRC scenarios arises from the diverse range of tasks, environments, and the varying capabilities that both humans and robots bring to collaborative efforts. Multiple studies have been conducted in an attempt to define the levels of interaction in human-robot collaboration processes \cite{aaltonen2018refining,badia2022virtual}. However, the majority of existing research predominantly addresses specific facets of human-robot interaction (HRI), potentially overlooking the broader span of interaction dynamics. A more flexible approach focusing on industrial settings is proposed by Mukherjee et al. in \cite{mukherjee2022survey}, where authors' classification criteria include the characteristics of the task and the workspace, autonomy level and operational mode of the robot, and the allowance for physical contact between the agents. This taxonomy ranges from level zero, being fully programmed robotics, to level five, classified as completely autonomous robots. Interestingly, both ends of the spectrum envisage no human interaction. Level four, the collaboration, involves humans and robots working simultaneously towards common goals. A more detailed and organized view of the approach from \cite{mukherjee2022survey} is shown in Table \ref{table:1}.

\begin{table}[ht!]
\caption{Industrial HRI levels, adopted from \cite{mukherjee2022survey}}
\label{table:1}
\begin{tabular}{ m{0.05\linewidth} p{0.18\linewidth} p{0.69\linewidth} }
\toprule
Level & Interaction      & Description                                                                                                                                                                     \\ \midrule
L0    & Fully Programmed & Traditional approach with physically restricting cages, no consideration of HRC.                                                                                                \\
L1    & Co-existence     & The agents are separated by safety zone, the robot pauses its operation in case human enters the area.                                                                          \\
L2    & Assistance       & Robot can assist human in a certain task (such as operations with heavy objects), but it has no independent tasks.                                                          \\
L3    & Co-operation     & The agents work together towards the common goal within the designated intervention zone. However, human and robot do not share the same task and there is no physical contact. \\
L4    & Collaboration    & Humans and robots working autonomously towards the same goal sharing the the task, workspace, and resources.                                                                                \\
L5    & Fully Autonomous & Generalizable manipulators trained using ML algorithms, no human intervention is considered.                                                                                    \\ \bottomrule
\end{tabular}
\end{table}

Moreover, we believe that the synergy between machine learning and extended reality presents a unique potential to provide an intuitive approach for human operators to act as robots' instructors. In this work, we propose an additional extension of the above-mentioned levels from \cite{mukherjee2022survey}, four and five, termed "fully autonomous with the human-in-the-loop." While this sub-level may not require a continuous human-robot interaction, it does offer the possibility for a human operator to step in during the autonomous process, augmenting it with human expertise as needed.

The goal of this paper is to conceptualize approaches for human involvement with the autonomous ML-based manipulator to facilitate the transfer of human expertise and skill. For this purpose, XR is employed as a communication middleware, thereby enhancing and simplifying the agent's interaction process. Additionally, the review of XR implementations in the current literature for HRC is presented, with a primary focus on manipulator arms in industrial settings. Furthermore, based on the presented conceptualization and conducted review, we show that integration of extended reality and machine learning could serve as the foundational pillar for the future of autonomous robotics and smart manufacturing in the context of Industry 5.0 and human-centricity.

\section{FRAMEWORK CONCEPTUALIZATION}
\label{section:framework}
Defining autonomous robotics is a critical aspect of this paper. Hence, Section \ref{subsection:task_framework} is dedicated to introducing the machine learning-based manipulator, and the respective framework for tasks generalization. Simultaneously, Section \ref{subsection:human_in_the_loop} delves into the integration of the human component through XR, specifically focusing on collaboration, programming, and performance oversight.

\begin{figure*}[b!]
    \centering
    \includegraphics[width=0.95\textwidth]{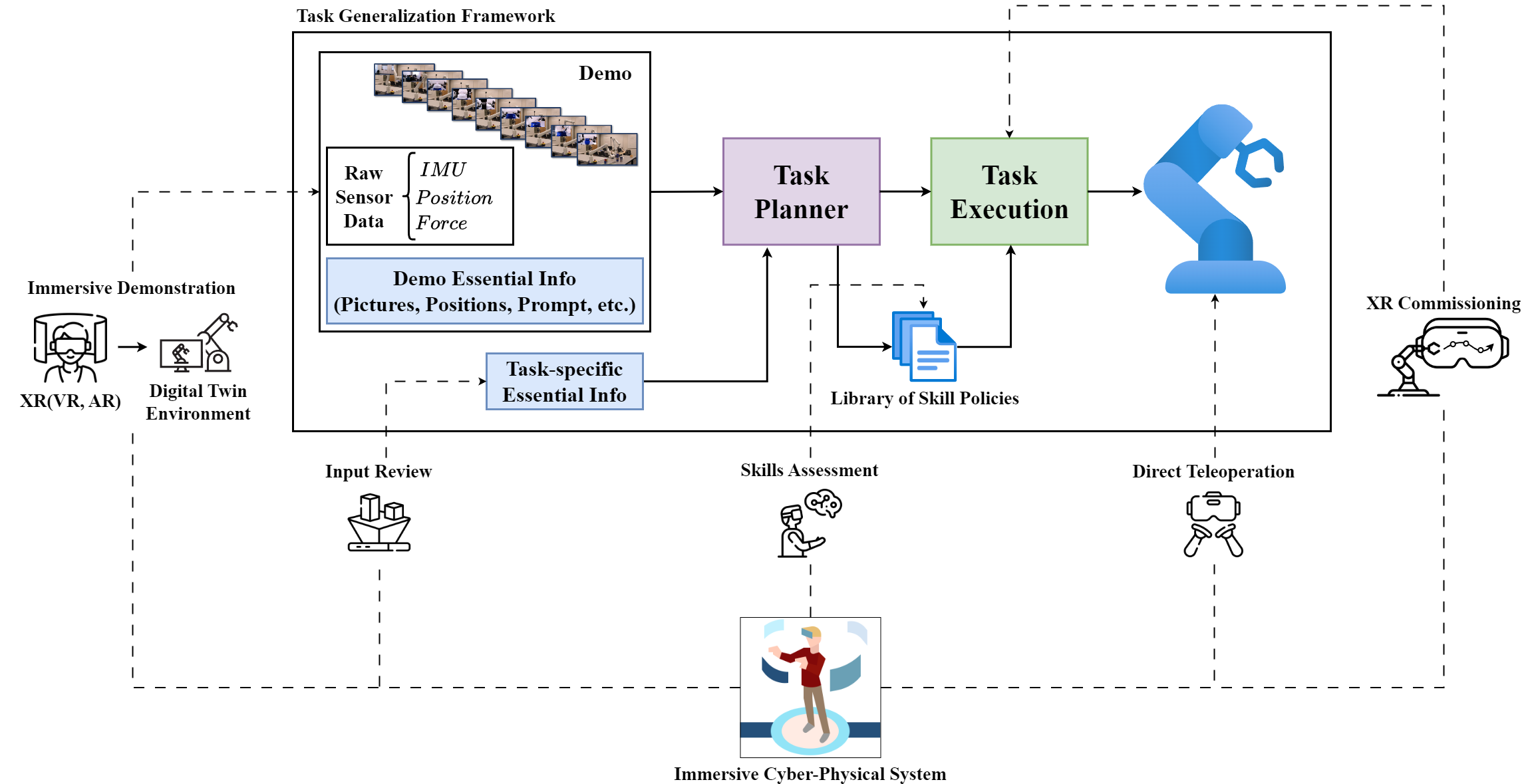}
    \caption{The outline of the proposed framework with Human-in-the-Loop}
    \label{fig:human_centric_framework}
\end{figure*}

\subsection{Manipulator Task Generalization}
\label{subsection:task_framework}
In this subsection, we provide a high-level overview of the ML-based autonomous manipulator, defining its autonomy and capacity for generalizable operations. In order to better understand the specific technologies and methods that enable this system, we recommend turning to appropriate literature: the authors in \cite{fang2012interactive} provide an overview of a similar system architecture using traditional control methods. Meanwhile, studies such as \cite{lin2022manipulation} and \cite{nasiriany2022augmenting} introduce ML-related components.

It is important to emphasize that the autonomous robot is equipped with pre-trained skill policies. Hence, the Task Generalization Framework (TGF), shown in the center of Fig. \ref{fig:human_centric_framework}, is designed to increase the manipulator's ability to generalize. The proposed task generalization framework for the manipulator is constructed of a hierarchy of modules and can be largely divided into three major sections: Demo, Task Planner, and Task Execution. 

\subsubsection{Demo}
To facilitate the robot's learning of a new task, a demonstration, also known as a sample task, is provided. This demonstration serves as a representation of the presumed task that the manipulator is expected to perform. The demo task, along with essential information about the actual task or the manipulated object, is provided to the next module, the Task Planner. Depending on the real scenario, the essential information may include numerical measurements, colours, shapes, CAD models, etc. 

\subsubsection{Task Planner}
It is assumed that the actual task executes similar skills compared to the demo task, but in a different environment (e.g., varied toolsets, obstacles, and parametric characteristics). Therefore, the Task Planner decomposes the provided demonstration (i.e., sample task) into multiple skills, which are then associated with and utilized for the execution of the actual task. Usually, skills are defined as low-level abstractions or primitives \cite{Saukkoriipi2020RobotSkill}; for example, the task is decomposed into the commands to ``move linearly'', ``locate the object'', ``position the gripper'', ``pick up the object'', etc. This decomposition forms the foundation of the skills library, a collection of fundamental actions essential for task completion. By providing crucial information, the skills library enhances the efficiency of task execution in different environments.

\subsubsection{Task Execution}
In the Task Execution module, the agent builds up a heterogeneous spatial representation to localize itself in the environment. Using the spatial representation, the information received from the Task Planner, and the pre-trained skill-specific policy, the trajectory planner maps the robot motion and carries out the instructed skill.

\subsection{Human-in-the-Loop Component}
\label{subsection:human_in_the_loop}
The use of immersive technology that augments or completely replaces the real world opens up opportunities for safe and intuitive human-robot interaction and task programming. We present XR-based interaction approaches with the goal of integrating human intelligence into the robot's learning process for task generalization. Also, the methods described below are easily translated to various pre-trained autonomous manipulators. We further elaborate on the conceptual basis for each method outlined in Fig. \ref{fig:human_centric_framework}.

\subsubsection{Immersive Demonstration}
In the proposed concept the human experts can deliver their skills through immersive demonstration, which is a convenient, intuitive, and safe way of illustrating the task at hand. The sample task can be demonstrated through a virtual environment, where all variables are controlled. In general, the immersive demonstration through VR can also be used as the sole source of training data, or it may complement previous demonstrations (whether in virtual or physical worlds) to increase the manipulator's ability to generalize and perform a larger set of tasks. Additionally, the virtual demonstration does not limit the operator in terms of the physical location. 

Another extension of this method is the on-site AR-based programming. The demonstration using AR, conducted on the manipulator's virtual model, can be interpreted as a virtual kinesthetic teaching \cite{iqbal2021exploring}. It does not have to interrupt the production process since the manipulations are applied to the virtual twin of the robot. 

\subsubsection{Input Review}
The opportunity to review the task-specific information ensures the accuracy and relevance of the given data, refining the overall process for optimal performance and adaptability in task execution. The exact implementation varies on the type of fed data; it may be visualized through AR or within VR space for the operator to manipulate and modify.

\subsubsection{Skill Assessment}
The breakdown and abstraction of skills can be revised based on real-time insights overseen by the human. The option to visualize (whether in AR or VR) and review the work done by the Task Planner module gives some level of transparency to the machine learning black box. The human operator can modify existing skills or incorporate new ones using XR as a virtual programming interface.

\subsubsection{Direct Teleoperation}
The opportunity to assume complete control of the manipulator through a VR headset enables humans to personally coordinate the robot's actions for task completion. This capability may be particularly useful in case the manipulator fails to accomplish the task repeatedly. Additionally, the requirement for the operator to be physically present on site is eliminated, as teleoperation can be carried out from a remote location. Furthermore, the experience gained through direct teleoperation could be utilized to adjust the manipulator's skill policies.

\subsubsection{XR Commissioning}
The ability to program the robot through the use of XR serves as a validation tool, granting control to a human for reviewing and modifying the visualized path and trajectory parameters. This approach, termed XR commissioning, proves useful during the commissioning of a new manipulator or when the operator intends to add or modify the existing programs. When the modification is implemented, essentially, it establishes a new ground truth in the retraining process of the robot's trajectory planner, integrating human experience into AI.

\subsection{Key Technologies}
\label{subsection:keyTech}
This subsection aims to give a broader context for how the advancements of particular technologies could influence the future of human-in-the-loop frameworks and human-robot collaboration in general.

\subsubsection{Extended Reality}
Through the simulation of intricate training scenarios, provision of augmented overlays for task guidance, and facilitation of remote collaboration, XR fosters a more fluid communication interface between humans and robots. This enhancement in communication offers a foundation for improvements in the precision, safety, and efficiency of applications within the field of HRC.

\subsubsection{Digital Twin}
The digital twin concept involves creating a dynamic digital representation of a physical system, enabling simulation, analysis, and optimization. In HRC, digital twins can be used to design and test collaborative processes, predict maintenance needs, and improve system adaptability \cite{feddoul2023exploring}. Furthermore, the digital twin is a key concept of cyber-physical systems that provides real-time control and monitoring, essential for certain aspects of XR.

\subsubsection{Artificial Intelligence}
Artificial Intelligence, leveraging machine learning algorithms and models, enhances task planning, decision-making, and environmental perception within HRC. This enables robots to intelligently interpret human gestures, speech, and intent while adapting to their surroundings for more strategic planning \cite{mukherjee2022survey}. Such AI-driven capabilities ensure seamless adaptation to human behaviors, deepening interaction and improving cooperation, making HRC systems capable of executing sophisticated, context-aware actions \cite{jahanmahin2022human}.

\subsubsection{Cloud Computing}
Cloud computing provides the infrastructure for scalable and on-demand computing resources, critical in managing the extensive data generated in industrial HRC settings. It enables the centralization of data analysis and storage, offering robust platforms for AI and machine learning models to operate efficiently. This technology strengthens the flexibility of HRC systems, enabling them to adapt to new tasks and environments quickly by leveraging cloud-based knowledge and computational power \cite{inamura2021sigverse}.

\subsubsection{Edge Computing}
Edge computing processes data near its source, reducing latency and enabling real-time responses critical for human-robot collaboration. By decentralizing computation, it ensures swift data analysis, essential for tasks needing immediate feedback. This enhances robotic autonomy, safety, and operational efficiency, especially in environments where split-second decisions are crucial \cite{shojaeinasab2022intelligent}. Edge computing's integration into HRC systems supports seamless operation and higher responsiveness, aligning with the demands of advanced manufacturing and collaborative tasks.

\section{APPLICATIONS REVIEW}
\label{section:applications}
This section aims to review recent studies related to XR application areas in HRC and complement the proposed human-in-the-loop framework with the existing practical implementations. The reviewed use cases are conditionally divided into four categories for a more organized and comprehensive representation, offering diverse viewpoints through the lenses of operator support and communication, safety considerations, teleoperation, and robot programming.

\subsection{Operator Support and Communication}
The idea of using extended reality for the purpose of operator support is not new, and it is not akin to purely HRC. Virtual and augmented reality have been applied in the areas of product development \cite{berg2017industry} or operator task training \cite{gavish2015evaluating}. The authors in \cite{papanastasiou2019towards} define the uses of XR for operator support in the following ways: a) show visual and text information regarding the process, b) provide the operator with visual and audio cues warning about certain dangers, such as the movement of the robot, c) visualize the area used by the robot within the real environment to minimize the risk of collisions.

Overall, the reviewed literature supports the classification by \cite{papanastasiou2019towards}. Bolano et al. \cite{bolano2021deploying} presented an interface that visualizes the swept volume of the robot's planned motion using a point cloud, which allowed the operator to foresee the volume that the robot will occupy. Chu et al. in \cite{chu2023augmented} proposed two AR-based visual interfaces to provide human operators with situational awareness. One of the interfaces displayed a semi-transparent barrier next to the manipulator, warning the operator of the robot's working envelope. Another interface displayed the virtual gripper model that moved along the robot's trajectory a few seconds before the physical robot, giving the operator enough time to assess the situation. The study by Dimitropoulos et al. \cite{dimitropoulos2021seamless} proposed a human-to-robot collaboration interface involving both AR technology and machine learning. The authors deploy a convolutional neural network to an AR headset to assist the operator in detecting the assembly parts of interest. Moreover, the authors position several markers throughout the testing environment, which are then detected by the AR headset, enabling the locating and tracking of operators in the scene, therefore eliminating the necessity for multiple stationary RGB cameras. This approach enhanced the flexibility of operator tracking and provided input data to the manipulator regarding the operator's actions and movements. Furthermore, for the collaborative tasks, the implemented interface included gesture-based commands letting the user modify the end-effector position if required.

It is important to highlight that the classification outlined in \cite{papanastasiou2019towards} primarily emphasized a passive approach, focusing on supporting functions and certain safety aspects for humans in proximity to the robot. Nonetheless, a more proactive perspective would involve considering human-to-robot communication, enabling the manipulator to find the optimal way to assist the human operator. In general, the utilization of extended reality head-mounted displays offers various input modes for controlling the manipulator, including speech, gaze, and hand gestures. Authors in various studies explored these modes individually, as well as the possibility of fusion. In \cite{bolano2021deploying}, the authors showed two separate communication interfaces allowing the operator to either use the voice command or point with the hand. Meanwhile, the authors in \cite{chen2022real} attempted to let the operators use both voice and hand gestures simultaneously. The proposed framework included the calculation of the confidence score for each communication mode to address controversial input data. Nonetheless, another methodology was investigated by Mukherjee et al. in \cite{mukherjee2022ai}, where authors proposed the AI-powered multi-modal fusion architecture based on fuzzy inference and Dempster-Shafer theory to deal with incomplete or conflicting evidence. The experiments were conducted using voice commands and hand gestures, however, authors claim the model should be sufficiently generalizable to include other modes of input.

\subsection{Safety Considerations}
\label{subsection:safety}
In human-robot collaboration scenarios, the operator's safety is a primary concern. Multiple frameworks have been developed based on monitoring separation, speed, power, and force limitations. Lately this study has been further extended by trying to predict possible collisions through the optimization-based control methods. The implementation of extended reality cannot directly solve the optimization-related problems of collision avoidance, but it may provide a flexible solution to increase the operators' safety.

Cogurcu et al. \cite{cogurcu2023augmentedCages} suggested an AR-based virtual safety zone system around the manipulator comparable with cell cages for industrial robots. The virtual barriers are dynamic, changing position relative to the manipulator movements. If a human enters the safety zone, the robot stops immediately. A similar but inverse approach has been taken by Hoang et al. \cite{Croft_2022_Virtual_Barriers}, guaranteeing an effective way to track human motion by creating a virtual barrier around the user anchored to the AR headset, allowing the person to move around the workplace freely. In case the robot detects an edge of the barrier on its planned path, it must adapt to avoid collision. The work goes even further and showcases the possibility of adding obstacle-oriented virtual barriers restricting the robot's motion in certain areas. The implementation in \cite{cogurcu2023augmentedCages} and \cite{Croft_2022_Virtual_Barriers} require the operator to wear the XR headset continuously. Potentially, this methodology could be used as a way to gather operator movement data in order to learn and manage the individual operators' preferences (related methodologies are also described in \cite{dimitropoulos2021seamless} and \cite{choi2022integrated}) or to use in the collision prediction as task-specific historical data.  

The authors in \cite{tsamis2021intuitive} utilize AR technology to visualize the robot's working envelope. Furthermore, the virtual twin of the physical robot is visualized to give the user a better idea of the planned manipulator motion and future positions. Meanwhile, \cite{choi2022integrated} proposed a significantly more complex architecture consisting of the robot's digital twin, deep learning model, two depth sensors, and the mixed reality headset. The authors investigate different strategies to extract data on the operator's location to synchronize it with the robot's digital twin. Essentially, the study manages to accurately calculate and visualize the distance between the operator's hands and the manipulator in real time, leading to better surroundings and safety awareness.

Some authors approach the concerns for safety from another corner of the XR paradigm - Virtual Reality. Creating a completely virtual environment allows the mimicking of realistic as well as potential hypothetical scenarios \cite{badia2022virtual}, where users can interact and familiarize themselves with the equipment at no risk of injury. Additionally, the concept of HRI in virtual reality can be elevated by incorporating the digital twin of the manipulator. Hence, it is no longer just a training simulation, but a real-time teleoperation (discussed in detail in \ref{subsection:teleoperation}) that blurs the boundary between the virtual and physical interaction \cite{kuts2022digital}. On the other hand, the aspects of mental and physical load of interactions in VR are not fully examined, and concerns of cybersickness should be addressed through further research \cite{souchet2023narrative}.

\subsection{Teleoperation}
\label{subsection:teleoperation}
Described as the remote, real-time control of the robot, teleoperation is a widely studied area of research in robotics \cite{solanes2020teleoperation}. The teleoperation process is usually associated with multiple challenges \cite{zein2021deep}. The first issue is mapping a rather large number of joints and degrees of freedom to the human's control interface. Secondly, poor perception leads to lower situational awareness, therefore influencing the overall ability and efficiency of the operator to accomplish the task. Finally, the task planning process for an operator from a remote location is particularly hard due to the need to breakdown the high-level objective into a low-level sequence of actions. Therefore, this section provides an overview of studies that address the above-mentioned issues by leveraging XR technology.

Kennel-Maushart et al. \cite{kennel2022multi} presented an MR interface for multi-robot systems that allows the operator to specify target poses, avoiding unfavorable setups that lead to singularities. The authors present their optimization method tested on a dual-arm ABB YuMi via the developed MR interface, allowing the user to teleoperate the payload in real-time and remotely. In order to adjust the orientation, position, velocity, and force of the robot, Sun et al. \cite{sun2020new} introduced an MR-based teleoperation interface with an integrated series of fuzzy-based algorithms, improving the overall maneuverability of the system. 

One of the most interesting sub-domains for research is multi-view teleoperation, where the operator has access to several points of view, solving problems of occlusions and leading to better spatial awareness. One of the most common and straightforward approaches is picture-in-picture (PIP), where multiple video streams are overlaid simultaneously. Usually, the global view is represented as the third-person view of the system, while the local view is extracted from the camera attached to the end-effector. The primary issue with the PIP method is the need for operators to frequently switch between views, resulting in a continuous change of operating perspectives. A multi-view fusion method is presented in \cite{wei2021multi} showcasing the possibility to construct a 3D point cloud reconstruction of the objects that are occluded in one of the views. The authors use a VR headset as the basis for their interface. The global view, captured by a stationary stereo camera, is displayed alongside the visual augmentations for the occluded objects (extracted from the local view). Furthermore, the occluded robot components, such as gripper fingers reaching for an object in the box, are also rendered as a visual augmentation. 

Kuts et al. in \cite{kuts2022digital} investigated the viability of the digital twin (DT) as the validation tool for industrial robot manipulation. The implemented framework includes the DT of the manipulator in the virtual environment, which is fully synchronized with the physical robot. The VR interface for robot control includes the possibility of changing the joint rotation angles, speed, and gripper function. This approach lets the end users remain in the decision loop remotely and in real time. Additionally, the presented interface in \cite{kuts2022digital} requires the operator to manually modify the position of each joint until reaching the destination, which, in fact, leads to the idea of the task-level authoring \cite{senft2021task} - forcing the human to break the objective into smaller steps. 

Meanwhile, DelPreto et al. \cite{delpreto2020helpingRobotsLearn} presented an online learning framework where human demonstrations are conducted in order to complement ML-based autonomous robots. The robot uses self-supervised learning, however, if the task cannot be properly accomplished, it request a direct demonstration from a human operator that is performed via Virtual Reality. The work in \cite{delpreto2020helpingRobotsLearn} is a great example of autonomous robotics with the human-in-the-loop, where XR acts as a human-robot communication middle-ware complementing AI algorithms with the human experience.

\subsection{Robot Programming}
\label{subsection:robotProgramming}

Robot programming, including operations such as relocation, grasping, and orientation change, are all among the most important functionalities of a robot \cite{li2023immersive}. In general, robot control methods can be divided into traditional and learning-based methods. The traditional control, delivered through offline programming, allows robot actions to be fully programmed. Nonetheless, it lacks the flexibility required in rapidly changing environments where it is nearly impossible to foresee all circumstances. Kinesthetic teaching partially addresses these concerns by enabling the user to easily and directly modify or build from scratch the robot's waypoints, grasping positions, etc. On the other hand, learning-based methods generally employ the use of machine learning algorithms. The implementation of AI opens opportunities for a larger degree of autonomy, better generalization in tasks, and even behavior modeling. 

In 2012, Fang et al. \cite{fang2012interactive} introduced an interactive framework based on Augmented Reality (AR) for adjusting a robot's path. The authors incorporated their framework with the robot's task and trajectory planner, enabling the operator to review the initial path. This integration offered the flexibility to modify, add, or delete waypoints between the starting and destination points. Quintero et al. \cite{quintero2018robot} presented a trajectory modification interface similar to the one in \cite{fang2012interactive}. However, the authors also conducted a study to compare it to kinesthetic teaching. The findings indicate that AR-based trajectory modification frameworks can reduce the teaching time, and show better overall performance since it is easy to use and is less physically demanding. 

Luebbers et al. \cite{luebbers2021arc} proposed a method of constrained learning from demonstration with the purpose of long-term skill maintenance of the manipulator. The introduced AR interface allows users to visualize and modify the task-associated gripper positions as well as the movement constraints. Interestingly, the idea of managing virtual constraints (also referred to as barriers) is similar in nature between \cite{luebbers2021arc}, where authors use it for robotics path planning to accomplish a task, and \cite{Croft_2022_Virtual_Barriers}, where the primary subject of interest is safety aspects.

The term HRC often implies an arm manipulator, however, HRC can also refer to other robotic platforms as well. In fact, many researchers attempt to study the interaction methods with mobile robots, including the role of XR in the process. For example, Tsamis et al. \cite{tsamis2021intuitive} presented an AR-based framework of a manipulator on a mobile platform. The mobile robot navigates to the goal pickup position, where the arm utilizes object detection to plan its path for grasping. AR plays a key role in keeping humans in the decision-making process by reviewing the planned routes of both agents. Focusing fully on mobile robotics, Gu et al. \cite{gu2022ar} presented a simple yet effective AR-based interface for navigation goals. The AR Point\&Click interface allows the use of natural pointing gestures, which are captured and interpreted by the cameras on the AR headset. The authors compare their approach to several other methods and conclude that based on user study, the proposed interface leads to higher efficiency and reduced mental load. Although the presented implementation in \cite{gu2022ar} was done for a mobile robot, it easily translates to the arm manipulator setting - a similar example is illustrated in \cite{yan2023complementary} as part of a larger research on MR interfaces for HRC.

As mentioned earlier in this subsection, the use of ML has become widespread in the robotics industry, and in this context, XR also establishes its relevance, particularly within the domain of imitation learning. One of the most famous works in this area was published in 2018 by Zhang et al. \cite{zhang2018deep}, showcasing a method to directly map pixels to actions from the demonstrations obtained in the virtual environment. Interestingly, an inexpensive system with less than 30 minutes of demonstration was sufficient to achieve nearly 90\% success rate. Similarly, Dyrstard et al. \cite{dyrstad2018grasping_fish} investigated the possibility of skill transfer for fish grasping tasks. By collecting just a few dozen demonstrations in virtual reality and employing domain randomization, a substantial synthetic training dataset comprising 100,000 samples was generated. Considering the given task and setting, the authors managed to achieve 74\% accuracy in grasping. After a more thorough analysis and dismissal of non-ML-related failures, the success rate could be estimated at 80\%.

\section{DISCUSSION}
\label{section:discussion}
The discussion section aims to bring out the major advantages and limitations and summarize the future research outlook for extended reality within industrial HRI.

\subsection{Mitigating Risks}
The visualization of motion intention and object manipulation gives the operator a better understanding of the workflow. In the case of VR systems, one of the main advantages is the elimination of the need for physical expert presence \cite{arents2022smart}. In general, studies show that XR is a unique tool that allows the conduct of teleoperation, robot programming, and various operator-supporting functions in a safe and controlled manner. Although XR offers the possibility of finding a balance between operator safety and robot efficiency, the impact on physical and mental health from working with head-mounted displays in HRI tasks necessitates further studies. As mentioned in \cite{souchet2023narrative}, there are currently no optimal solutions to address all possible side effects like muscle fatigue, motion sickness, and mental overload. Similarly, for human-robot collaborative tasks, the psychological factor remains a significant area of research. This includes the effort to cultivate trust between the agents, as well as methods to generate motion and trajectories that closely mimic human behavior \cite{hong2023humanrobot}.

\subsection{Immersiveness}
 One of the bases for introducing XR is its immersive potential. The immersive environment allows users to perceive the spatial aspects of a robot and its surroundings more effectively. By providing a realistic and engaging experience, users can interact with and understand the robot's movements and actions in a way that is not possible through traditional interfaces. This level of interaction and understanding can facilitate better demonstration quality, which is one of the most decisive factors in the effectiveness of robot policy learning \cite{li2023immersive, jackson2019benefits}.

Given the enhanced visualization and situational awareness provided by immersive capabilities, task coordination is another area that can be improved for better collaboration between agents. The application of XR in the context of multi-robot systems can provide easy-to-use, intuitive methods for commanding multiple agents synchronously \cite{kennel2023interacting}.

\subsection{User-oriented Concepts}
In order to best address human-robot collaboration, it is essential to take social cues into account. That can be expressed in terms of communication modes like gestures, voice, or gaze - all of which are supported by the modern XR headsets. The use of head-mounted displays (HMDs) eliminates the need for multiple stationary cameras and various additional sensors. However, further investigation is needed on how to capture and accurately interpret multi-modal communication signals, such as fusion techniques. Similarly, the tracking functionality in HMDs opens certain possibilities to study operator preferences. The authors in \cite{nemlekar2023transfer} propose a method to transfer operator preferences from a canonical to an actual assembly task, allowing the cobot to assist the operator proactively. The incorporation of XR in this process could facilitate data collection - operator and surrounding related, and potentially result in a more personalized HRI experience.

\subsection{From Lab to Industry}
One of the primary obstacles preventing the adaptation of extended reality for the human-robot collaboration scenarios is setup costs, requiring substantial initial investment. Additionally, there is a lack of unified consensus within industry on the XR integration strategy and interface development, which understandably stems from the differences in industry-specific requirements and individual products. A possible future area of research could involve investigating the feasibility of utilizing the same XR interface across multiple manipulators with the flexibility to easily customize the interface for new robot- or product-specific characteristics. Furthermore, the implementation of XR is usually performed in a bundle with other digital technologies, as outlined in \ref{subsection:keyTech}. Therefore, it entails additional investments and resources, and in fact, the effectiveness of XR becomes conditional on the development and maintenance of other technologies. Also, it is noteworthy that most of the reviewed works perform experiments in the laboratory. The prospect of transferring the developed methodologies and interfaces to real-world scenarios and industrial settings remains yet to be explored.

\section{CONCLUSIONS}
\label{section:conclusion}
This study investigates the application of extended reality to enhance human-to-robot interaction with a focus on the industrial setting. The conducted review suggests that recent advancements in extended reality make it a practical communication interface for HRC, particularly when integrated with other enabling technologies such as digital twinning and machine learning. The presented conceptualization of the framework for \textit{fully autonomous manipulator with the human-in-the-loop} could be considered as a stepping stone towards more effective human-robot collaboration as it aims to strike the balance between autonomy, efficiency, and operational flexibility. Overall, the incorporation of immersive technology augments human control over both the robot's movements and the surrounding environment, leading to further adaptation of human-centric cyber-physical systems.


\section*{ACKNOWLEDGMENT}
We would like to acknowledge the financial support of the Natural Sciences and Engineering Research Council (NSERC) Canada under the Discovery Grant RGPIN-2023-05408 in this research.


\bibliographystyle{ieeetr}
\bibliography{references.bib}

\begin{thebibliography}{10}

\bibitem{jahanmahin2022human}
R.~Jahanmahin, S.~Masoud, J.~Rickli, and A.~Djuric, ``Human-robot interactions in manufacturing: A survey of human behavior modeling,'' {\em Robotics and Computer-Integrated Manufacturing}, vol.~78, p.~102404, 2022.

\bibitem{pizzagalli2021ucd}
S.~Pizzagalli, V.~Kuts, and T.~Otto, ``User-centered design for human-robot collaboration systems,'' in {\em IOP Conference Series: Materials Science and Engineering}, p.~012011, IOP Publishing, 2021.

\bibitem{dianatfar2021review}
M.~Dianatfar, J.~Latokartano, and M.~Lanz, ``Review on existing vr/ar solutions in human--robot collaboration,'' {\em Procedia CIRP}, vol.~97, pp.~407--411, 2021.

\bibitem{suzuki2022augmented}
R.~Suzuki, A.~Karim, T.~Xia, H.~Hedayati, and N.~Marquardt, ``Augmented reality and robotics: A survey and taxonomy for ar-enhanced human-robot interaction and robotic interfaces,'' in {\em Proceedings of the 2022 CHI Conference on Human Factors in Computing Systems}, pp.~1--33, 2022.

\bibitem{speicher2019mixed}
M.~Speicher, B.~D. Hall, and M.~Nebeling, ``What is mixed reality?,'' in {\em Proceedings of the 2019 CHI conference on human factors in computing systems}, pp.~1--15, 2019.

\bibitem{aaltonen2018refining}
I.~Aaltonen, T.~Salmi, and I.~Marstio, ``Refining levels of collaboration to support the design and evaluation of human-robot interaction in the manufacturing industry,'' {\em Procedia CIRP}, vol.~72, pp.~93--98, 2018.

\bibitem{badia2022virtual}
S.~B.~i. Badia, P.~A. Silva, D.~Branco, A.~Pinto, C.~Carvalho, P.~Menezes, J.~Almeida, and A.~Pilacinski, ``Virtual reality for safe testing and development in collaborative robotics: challenges and perspectives,'' {\em Electronics}, vol.~11, no.~11, p.~1726, 2022.

\bibitem{mukherjee2022survey}
D.~Mukherjee, K.~Gupta, L.~H. Chang, and H.~Najjaran, ``A survey of robot learning strategies for human-robot collaboration in industrial settings,'' {\em Robotics and Computer-Integrated Manufacturing}, vol.~73, p.~102231, 2022.

\bibitem{fang2012interactive}
H.~Fang, S.~Ong, and A.~Nee, ``Interactive robot trajectory planning and simulation using augmented reality,'' {\em Robotics and Computer-Integrated Manufacturing}, vol.~28, no.~2, pp.~227--237, 2012.

\bibitem{lin2022manipulation}
N.~Lin, Y.~Li, K.~Tang, Y.~Zhu, X.~Zhang, R.~Wang, J.~Ji, X.~Chen, and X.~Zhang, ``Manipulation planning from demonstration via goal-conditioned prior action primitive decomposition and alignment,'' {\em IEEE Robotics and Automation Letters}, vol.~7, no.~2, pp.~1387--1394, 2022.

\bibitem{nasiriany2022augmenting}
S.~Nasiriany, H.~Liu, and Y.~Zhu, ``Augmenting reinforcement learning with behavior primitives for diverse manipulation tasks,'' in {\em 2022 International Conference on Robotics and Automation (ICRA)}, pp.~7477--7484, IEEE, 2022.

\bibitem{Saukkoriipi2020RobotSkill}
J.~Saukkoriipi, T.~Heikkilä, J.~M. Ahola, T.~Seppälä, and P.~Isto, ``Programming and control for skill-based robots,'' {\em Open Engineering}, vol.~10, no.~1, pp.~368--376, 2020.

\bibitem{iqbal2021exploring}
M.~Z. Iqbal, E.~Mangina, and A.~G. Campbell, ``Exploring the real-time touchless hand interaction and intelligent agents in augmented reality learning applications,'' in {\em 2021 7th International Conference of the Immersive Learning Research Network (iLRN)}, pp.~1--8, IEEE, 2021.

\bibitem{feddoul2023exploring}
Y.~Feddoul, N.~Ragot, F.~Duval, V.~Havard, D.~Baudry, and A.~Assila, ``Exploring human-machine collaboration in industry: A systematic literature review of digital twin and robotics interfaced with extended reality technologies,'' {\em The International Journal of Advanced Manufacturing Technology}, vol.~129, no.~5, pp.~1917--1932, 2023.

\bibitem{inamura2021sigverse}
T.~Inamura and Y.~Mizuchi, ``Sigverse: A cloud-based vr platform for research on multimodal human-robot interaction,'' {\em Frontiers in Robotics and AI}, vol.~8, p.~549360, 2021.

\bibitem{shojaeinasab2022intelligent}
A.~Shojaeinasab, T.~Charter, M.~Jalayer, M.~Khadivi, O.~Ogunfowora, N.~Raiyani, M.~Yaghoubi, and H.~Najjaran, ``Intelligent manufacturing execution systems: A systematic review,'' {\em Journal of Manufacturing Systems}, vol.~62, pp.~503--522, 2022.

\bibitem{berg2017industry}
L.~P. Berg and J.~M. Vance, ``Industry use of virtual reality in product design and manufacturing: a survey,'' {\em Virtual reality}, vol.~21, pp.~1--17, 2017.

\bibitem{gavish2015evaluating}
N.~Gavish, T.~Guti{\'e}rrez, S.~Webel, J.~Rodr{\'\i}guez, M.~Peveri, U.~Bockholt, and F.~Tecchia, ``Evaluating virtual reality and augmented reality training for industrial maintenance and assembly tasks,'' {\em Interactive Learning Environments}, vol.~23, no.~6, pp.~778--798, 2015.

\bibitem{papanastasiou2019towards}
S.~Papanastasiou, N.~Kousi, P.~Karagiannis, C.~Gkournelos, A.~Papavasileiou, K.~Dimoulas, K.~Baris, S.~Koukas, G.~Michalos, and S.~Makris, ``Towards seamless human robot collaboration: integrating multimodal interaction,'' {\em The International Journal of Advanced Manufacturing Technology}, vol.~105, pp.~3881--3897, 2019.

\bibitem{bolano2021deploying}
G.~Bolano, Y.~Fu, A.~Roennau, and R.~Dillmann, ``Deploying multi-modal communication using augmented reality in a shared workspace,'' in {\em 2021 18th International Conference on Ubiquitous Robots (UR)}, pp.~302--307, IEEE, 2021.

\bibitem{chu2023augmented}
C.-H. Chu and Y.-L. Liu, ``Augmented reality user interface design and experimental evaluation for human-robot collaborative assembly,'' {\em Journal of Manufacturing Systems}, vol.~68, pp.~313--324, 2023.

\bibitem{dimitropoulos2021seamless}
N.~Dimitropoulos, T.~Togias, N.~Zacharaki, G.~Michalos, and S.~Makris, ``Seamless human--robot collaborative assembly using artificial intelligence and wearable devices,'' {\em Applied Sciences}, vol.~11, no.~12, p.~5699, 2021.

\bibitem{chen2022real}
H.~Chen, M.~C. Leu, and Z.~Yin, ``Real-time multi-modal human--robot collaboration using gestures and speech,'' {\em Journal of Manufacturing Science and Engineering}, vol.~144, no.~10, p.~101007, 2022.

\bibitem{mukherjee2022ai}
D.~Mukherjee, K.~Gupta, and H.~Najjaran, ``An ai-powered hierarchical communication framework for robust human-robot collaboration in industrial settings,'' in {\em 2022 31st IEEE International Conference on Robot and Human Interactive Communication (RO-MAN)}, pp.~1321--1326, IEEE, 2022.

\bibitem{cogurcu2023augmentedCages}
Y.~E. Cogurcu and S.~Maddock, ``Augmented reality safety zone configurations in human-robot collaboration: A user study,'' in {\em Companion of the 2023 ACM/IEEE International Conference on Human-Robot Interaction}, pp.~360--363, 2023.

\bibitem{Croft_2022_Virtual_Barriers}
K.~C. Hoang, W.~P. Chan, S.~Lay, A.~Cosgun, and E.~Croft, ``Virtual barriers in augmented reality for safe and effective human-robot cooperation in manufacturing,'' in {\em 2022 31st IEEE International Conference on Robot and Human Interactive Communication (RO-MAN)}, pp.~1174--1180, 2022.

\bibitem{choi2022integrated}
S.~H. Choi, K.-B. Park, D.~H. Roh, J.~Y. Lee, M.~Mohammed, Y.~Ghasemi, and H.~Jeong, ``An integrated mixed reality system for safety-aware human-robot collaboration using deep learning and digital twin generation,'' {\em Robotics and Computer-Integrated Manufacturing}, vol.~73, p.~102258, 2022.

\bibitem{tsamis2021intuitive}
G.~Tsamis, G.~Chantziaras, D.~Giakoumis, I.~Kostavelis, A.~Kargakos, A.~Tsakiris, and D.~Tzovaras, ``Intuitive and safe interaction in multi-user human robot collaboration environments through augmented reality displays,'' in {\em 2021 30th IEEE international conference on robot \& human interactive communication (RO-MAN)}, pp.~520--526, IEEE, 2021.

\bibitem{kuts2022digital}
V.~Kuts, J.~A. Marvel, M.~Aksu, S.~L. Pizzagalli, M.~Sarkans, Y.~Bondarenko, and T.~Otto, ``Digital twin as industrial robots manipulation validation tool,'' {\em Robotics}, vol.~11, no.~5, p.~113, 2022.

\bibitem{souchet2023narrative}
A.~D. Souchet, D.~Lourdeaux, A.~Pagani, and L.~Rebenitsch, ``A narrative review of immersive virtual reality’s ergonomics and risks at the workplace: cybersickness, visual fatigue, muscular fatigue, acute stress, and mental overload,'' {\em Virtual Reality}, vol.~27, no.~1, pp.~19--50, 2023.

\bibitem{solanes2020teleoperation}
J.~E. Solanes, A.~Mu{\~n}oz, L.~Gracia, A.~Mart{\'\i}, V.~Girb{\'e}s-Juan, and J.~Tornero, ``Teleoperation of industrial robot manipulators based on augmented reality,'' {\em The International Journal of Advanced Manufacturing Technology}, vol.~111, pp.~1077--1097, 2020.

\bibitem{zein2021deep}
M.~K. Zein, M.~Al~Aawar, D.~Asmar, and I.~H. Elhajj, ``Deep learning and mixed reality to autocomplete teleoperation,'' in {\em 2021 IEEE International Conference on Robotics and Automation (ICRA)}, pp.~4523--4529, IEEE, 2021.

\bibitem{kennel2022multi}
F.~Kennel-Maushart, R.~Poranne, and S.~Coros, ``Multi-arm payload manipulation via mixed reality,'' in {\em 2022 International Conference on Robotics and Automation (ICRA)}, pp.~11251--11257, IEEE, 2022.

\bibitem{sun2020new}
D.~Sun, A.~Kiselev, Q.~Liao, T.~Stoyanov, and A.~Loutfi, ``A new mixed-reality-based teleoperation system for telepresence and maneuverability enhancement,'' {\em IEEE Transactions on Human-Machine Systems}, vol.~50, no.~1, pp.~55--67, 2020.

\bibitem{wei2021multi}
D.~Wei, B.~Huang, and Q.~Li, ``Multi-view merging for robot teleoperation with virtual reality,'' {\em IEEE Robotics and Automation Letters}, vol.~6, no.~4, pp.~8537--8544, 2021.

\bibitem{senft2021task}
E.~Senft, M.~Hagenow, K.~Welsh, R.~Radwin, M.~Zinn, M.~Gleicher, and B.~Mutlu, ``Task-level authoring for remote robot teleoperation,'' {\em Frontiers in Robotics and AI}, vol.~8, p.~707149, 2021.

\bibitem{delpreto2020helpingRobotsLearn}
J.~DelPreto, J.~I. Lipton, L.~Sanneman, A.~J. Fay, C.~Fourie, C.~Choi, and D.~Rus, ``Helping robots learn: a human-robot master-apprentice model using demonstrations via virtual reality teleoperation,'' in {\em 2020 IEEE International Conference on Robotics and Automation (ICRA)}, pp.~10226--10233, IEEE, 2020.

\bibitem{li2023immersive}
K.~Li, D.~Chappell, and N.~Rojas, ``Immersive demonstrations are the key to imitation learning,'' in {\em 2023 IEEE International Conference on Robotics and Automation (ICRA)}, pp.~5071--5077, 2023.

\bibitem{quintero2018robot}
C.~P. Quintero, S.~Li, M.~K. Pan, W.~P. Chan, H.~M. Van~der Loos, and E.~Croft, ``Robot programming through augmented trajectories in augmented reality,'' in {\em 2018 IEEE/RSJ International Conference on Intelligent Robots and Systems (IROS)}, pp.~1838--1844, IEEE, 2018.

\bibitem{luebbers2021arc}
M.~B. Luebbers, C.~Brooks, C.~L. Mueller, D.~Szafir, and B.~Hayes, ``Arc-lfd: Using augmented reality for interactive long-term robot skill maintenance via constrained learning from demonstration,'' in {\em 2021 IEEE International Conference on Robotics and Automation (ICRA)}, pp.~3794--3800, IEEE, 2021.

\bibitem{gu2022ar}
M.~Gu, E.~Croft, and A.~Cosgun, ``Ar point \&click: An interface for setting robot navigation goals,'' in {\em International Conference on Social Robotics}, pp.~38--49, Springer, 2022.

\bibitem{yan2023complementary}
X.~Yan, Y.~Jiang, C.~Chen, L.~Gong, M.~Ge, T.~Zhang, and X.~Li, ``A complementary framework for human--robot collaboration with a mixed ar--haptic interface,'' {\em IEEE Transactions on Control Systems Technology}, 2023.

\bibitem{zhang2018deep}
T.~Zhang, Z.~McCarthy, O.~Jow, D.~Lee, X.~Chen, K.~Goldberg, and P.~Abbeel, ``Deep imitation learning for complex manipulation tasks from virtual reality teleoperation,'' in {\em 2018 IEEE International Conference on Robotics and Automation (ICRA)}, pp.~5628--5635, IEEE, 2018.

\bibitem{dyrstad2018grasping_fish}
J.~S. Dyrstad, E.~R. {\O}ye, A.~Stahl, and J.~R. Mathiassen, ``Teaching a robot to grasp real fish by imitation learning from a human supervisor in virtual reality,'' in {\em 2018 IEEE/RSJ International Conference on Intelligent Robots and Systems (IROS)}, pp.~7185--7192, IEEE, 2018.

\bibitem{arents2022smart}
J.~Arents and M.~Greitans, ``Smart industrial robot control trends, challenges and opportunities within manufacturing,'' {\em Applied Sciences}, vol.~12, no.~2, p.~937, 2022.

\bibitem{hong2023humanrobot}
J.~Hong, Z.~Zhang, A.~M.~S. Enayati, and H.~Najjaran, ``Human-robot skill transfer with enhanced compliance via dynamic movement primitives,'' 2023.

\bibitem{jackson2019benefits}
A.~Jackson, B.~D. Northcutt, and G.~Sukthankar, ``The benefits of immersive demonstrations for teaching robots,'' in {\em 2019 14th ACM/IEEE International Conference on Human-Robot Interaction (HRI)}, pp.~326--334, IEEE, 2019.

\bibitem{kennel2023interacting}
F.~Kennel-Maushart, R.~Poranne, and S.~Coros, ``Interacting with multi-robot systems via mixed reality,'' in {\em 2023 IEEE International Conference on Robotics and Automation (ICRA)}, pp.~11633--11639, IEEE, 2023.

\bibitem{nemlekar2023transfer}
H.~Nemlekar, N.~Dhanaraj, A.~Guan, S.~K. Gupta, and S.~Nikolaidis, ``Transfer learning of human preferences for proactive robot assistance in assembly tasks,'' in {\em Proceedings of the 2023 ACM/IEEE International Conference on Human-Robot Interaction}, pp.~575--583, 2023.

\end{thebibliography}

\end{document}